\documentclass[10pt,journal,compsoc]{IEEEtran}
\ifCLASSOPTIONcompsoc
\usepackage[nocompress]{cite}
\else
\usepackage{cite}
\fi
\ifCLASSINFOpdf
\else
\fi
\usepackage{verbatim} 
\usepackage{caption}
\usepackage{subcaption}
\usepackage{xcolor}
\usepackage{algorithmic}
\usepackage[linesnumbered,ruled,vlined]{algorithm2e}
\usepackage{amsmath,amssymb,amsfonts}
\usepackage{relsize}
\usepackage{graphicx}
\usepackage{textcomp}
\graphicspath{{./figures/}}
\usepackage{ragged2e}

\begin{document}
	%
	\title{Data Streaming and Traffic Gathering in Mesh-based NoC for Deep Neural Network Acceleration}
	%
	%
	%
	%
	
	\author{Binayak Tiwari, Mei Yang, Xiaohang Wang and Yingtao Jiang
		\IEEEcompsocitemizethanks{\IEEEcompsocthanksitem B. Tiwari, M. Yang, and Y. Jiang are with the Department
			of Electrical and Computer Engineering, University of Nevada, Las Vegas, NV  89154.\protect\\
			E-mail: btiwari@unlv.nevada.edu; mei.yang@unlv.edu; yingtao.jiang@unlv.edu
			\IEEEcompsocthanksitem Xiaohang Wang is with the Southern China Institute of Technology. E-mail: xh.wang@giat.ac.cn}
		\thanks{}}

	\IEEEtitleabstractindextext{%
		\justify
		\begin{abstract}

			The increasing popularity of deep neural network (DNN) applications demands high computing power and efficient hardware accelerator architecture. DNN accelerators use a large number of processing elements (PEs) and on-chip memory for storing weights and other parameters. As the communication backbone of a DNN accelerator, networks-on-chip (NoC) play an important role in supporting various dataflow patterns and enabling processing with communication parallelism in a DNN accelerator. However, the widely used mesh-based NoC architectures inherently cannot support the efficient one-to-many and many-to-one traffic largely existing in DNN workloads. In this paper, we propose a modified mesh architecture with a one-way/two-way streaming bus to speedup one-to-many (multicast) traffic, and the use of gather packets to support many-to-one (gather) traffic. The analysis of the runtime latency of a convolutional layer shows that the two-way streaming architecture achieves better improvement than the one-way streaming architecture for an Output Stationary (OS) dataflow architecture. The simulation results demonstrate that the gather packets can help to reduce the runtime latency up to 1.8 times and network power consumption up to 1.7 times, compared with the repetitive unicast method on modified mesh architectures supporting two-way streaming.

		\end{abstract}
		
		}

	\maketitle

	\IEEEdisplaynontitleabstractindextext

	%
	\IEEEpeerreviewmaketitle

\begin{figure}
		\centering
		\includegraphics[width=0.8\linewidth]{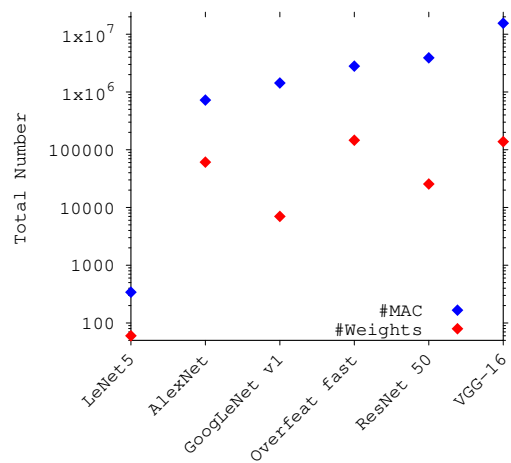}
		\caption{Some popular DNN models with number of weights and MAC operations.}
		\label{fig-numMAC}
	\end{figure}

	\IEEEraisesectionheading{\section{Introduction}\label{sec:introduction}}
	\IEEEPARstart{D}{eep} Neural Networks (DNNs) are widely adopted for a variety of applications, ranging from speech recognition, object detection and self-driving cars, to cancer detection, drug discovery, and genomics \cite{deeplearning-nature} \cite{selfdriving} \cite{cancerdetection}. DNNs are able to extract high-level features from input data, as in statistical learning, compared to hand-picked features from classic machine learning. This has enabled DNNs to achieve human-level accuracy, which comes at the cost of high communication and computation complexity. High complexity in DNNs is simply attributed to the huge number of parameters and multiply-and-accumulate (MAC) operations. Fig. \ref{fig-numMAC} shows the number of weights and MAC operations used in some of the popular DNN models. AlexNet \cite{alexnet} consists of 61M weights and 724M MACs, while VGG-16 \cite{vgg16} consists of 138M weights and 15.5G MACs.  
	
	\par 
	DNNs include the training phase and the inference phase. In the training phase, learning is involved in determining the network weights and the biases. The inference phase is actually the process of taking the inputs from the user or sensor and make use of the weights and biases obtained during the training phase to get the estimated result. Training DNNs often requires the use of a large dataset and is more computation-intensive than inference. Training is also a time-consuming process that may take up to several weeks at a cloud/data center. On the other hand, an inference can be employed even on edge devices like mobile phones. Due to the involvement of a huge amount of parameters (weights and biases Fig. \ref{fig-numMAC}), it is not possible to store all of these parameters in the local memory of the processing elements (PEs). Different levels of memory, from DRAM with high access cost to register files with low access cost \cite{hardwareforML}, are commonly used in DNN accelerators, which imposes the challenge of optimizing data movement in the memory hierarchy.
	\par
	In a DNN accelerator, PEs perform MAC operations, while the involved parameters are usually stored in the global memory. There is a need to transfer data from global memory to PEs, and vice versa. PEs and the memory elements are often interconnected by a Networks-on-Chip (NoC) \cite{noc-routepacketnotchips} \cite{maeri} \cite{spikingNN} to realize high throughput. These PEs operate in parallel and reduce the memory access as much as possible by sharing and reusing the parameters with each other, especially in spatial architectures. 
	\par
	As the communication backbone \cite{kilocore} \cite{all-to-all} \cite{pdp2020} of a DNN accelerator, NoC play an important role in supporting various traffic patterns and dataflow models, enabling processing with communication parallelism and enhancing scalability. NoC also offer a modular design property that helps in power gating. Mesh is a widely adopted NoC topology that is scalable and can support a large number of PEs. The artificial intelligence (AI) computing system from Cerebras \cite{cerebras} uses 2D mesh as a communication topology to connect thousands of AI cores. Groq \cite{groq} reorganizes 2D mesh into functional slices to optimize the microarchitecture. Existing mesh-based accelerator systems focus more on improving scalability and data reuse, and a little attention is given to enhancing communication support. Hence, this study is focused on efficient communication support for mesh-based DNN accelerators.
	\par
	Observing the nature of data traffic in DNN processing, there are many inputs/weights transferred from the global memory to PEs and results (like partial sums) collected from PEs to the global memory. This traffic can be classified as one-to-many (multicast) and many-to-one (gather) traffic, respectively. In the literature, different approaches have been proposed to support multicast traffic in NoCs \cite{pdp2020} \cite{surveymulticast}. Noticeably, in a DNN workload, the multicast traffic tends to have a fixed communication pattern. Thus, existing multicast algorithms may not be suitable for DNN workloads. In addition, support of gather traffic has not been well addressed in NoCs, as many-to-one traffic rarely occurs in conventional parallel workloads.     
	\par
	In this work, based on the observation that the multicast traffic in DNN workloads has the same source and destination set most of the time, we propose a modified mesh architecture with a one-way/two-way streaming bus to speedup the input and weight distribution in an NoC. In support of the many-to-one type of traffic on a mesh-based NoC, gather packets are used to efficiently deliver the partial sum results to the global memory. Streaming architectures and gather packets can be used on different dataflow models. To evaluate the performance of the proposed modified mesh-based architecture with gather support, analysis and simulations are conducted on the output stationary dataflow model using Alexnet \cite{alexnet} and VGG-16 \cite{vgg16} as a DNN workload, and compared with the conventional repetitive unicast method. 
	\par
	The remainder of the paper is organized as follows. Section \ref{sec:relatedwork} reviews dataflow models, NoC architectures, and communication support proposed for DNN accelerators. Section \ref{sec:motivation} provides the background and motivation behind this work. In Section \ref{sec:gather}, we describe the proposed methods and analyze data streaming architecture and gather support improvement. Section \ref{sec:evaluation} presents the experimental results of the proposed methods. Section \ref{sec:conclusion} concludes the paper.  

	\section{Related Work}\label{sec:relatedwork}
	DNN processing involves tens of layers and a large number of MAC operations using millions of parameters, which imposes tremendous throughput and energy-efficiency challenges to the computing platforms. Recently, spatial DNN accelerators like \cite{tpu} \cite{eyeriss} \cite{shidiannao} are gaining attention, as they are optimized to handle DNN processing effectively. Commonly used in FPGA and ASIC based designs \cite{hardwareforML} \cite{tpu} \cite{fpgadnn} \cite{binarizedNN}, spatial architectures use a distributed approach, which adopts a large number of PEs, each having its own control logic and limited local memory and shared global memory. Communication between PEs is allowed, which enables the data movement between them. In the design of a DNN accelerator, a major consideration is optimizing data movement, which aims to minimize the global memory access, and thereby reduce the power consumption during DNN processing.
	\par
	The dataflow determines the processing order, as well as where data is stored and reused, i.e., the way how data (i.e., inputs, weights, and partial sums) communication happens between the PE and memory element. In the literature, various dataflow models have been proposed \cite{hardwareforML}, including Weight Stationary (WS), Output Stationary (OS), Row Stationary(RS), and No Local Reuse (NLR); each of them has its own memory usage and energy advantages. In the WS model \cite{tpu}, weights are stationary at the PEs, while the inputs and partial sums move through the PEs and the memory element. On the contrary, the OS model has the output stationary at the PEs, while the inputs and weights move \cite{shidiannao}. The NLR model \cite{diannao} focuses on increasing the size of the global buffer at the expense of a register file, and thus decrease the DRAM accesses. The RS model increases the reuse of all data types rather than focusing on the reuse of one type. 
	\par
	DNN workloads contain different types of communication traffic that manages the data movement, such as partial sums, weights, and inputs streams to and from the memory. In a DNN accelerator \cite{dadiannao}, data movement is expensive in terms of energy, consuming around 50\% of the total energy. In some cases, data movement can even increase the latency \cite{eyeriss} due to the communication bottleneck. Although a bus-only based system has been proposed in some prior work, this kind of system will quickly become a bottleneck when the DNN size increases \cite{nocnn}. This observation leads to the works focused on the NoC architecture, and communication support of DNN accelerators \cite{maeri} \cite{spinnaker} \cite{neuronlink} \cite{closnn} \cite{socc} \cite{neu-noc}.
	\par
	Various studies have been done in NoC topology to accelerate a DNN workload \cite{dadiannao} \cite{nocnn} \cite{neu-noc} \cite{3dNN-acc}. In \cite{neu-noc}, a hierarchical Neu-NoC architecture that adopts a hybrid ring-mesh topology. Multiple PEs are connected in a group of rings connected via a mesh topology is proposed. This structure reduces the communication distance and shows better performance against the bus and tree structures. Other research \cite{3dNN-acc} propose, a reconfigurable topology for a 3D neural network accelerator that can be reconfigured as a tree to handle the multicast traffic. A many-core system SpiNNaker is proposed to simulate spiking neural networks with torus network topology. In \cite{nocnn}, the study looks at different topologies and concludes that the mesh NoC is better for realizing spiking neural networks, compared with the tree, point-to-point, and bus-based structures. In \cite{dadiannao}, a fat tree and a mesh are used for intrachip communication and data movement among chips, respectively. The separation of intrachip and interchip communication may create a bottleneck for the gather traffic abundant in a DNN workload.
	\par
	Changes in the NoC topology also cause a change in the communication cost and support of different traffic patterns. Another study \cite{eyeriss} proposes a mesh-based interconnection network called a hierarchical mesh network for DNN processing. The PEs and memory elements are grouped into a cluster, which is then connected via the hierarchical mesh network. The NoC is capable of configuring the network topology based on the needs. The NeuronLink \cite{neuronlink} is a chip-to-chip interconnection network for large neural networks that support both interchip and intrachip communication. Each chip consists of 16 PEs in a mesh topology, and 4 such chips are connected in a star topology to handle a large amount of unicast and multicast traffic.
	\par
	Various routing methods are adopted to fulfill the communication needs, especially multicast and gather traffic, in a DNN accelerator. Research in \cite{neuronlink} adopts XY routing for unicast traffic and a table-based routing for multicast traffic. However, another study \cite{eyeriss} proposes different NoC configurations for each datatype i.e, input activation, weights, and partial sums. Further, this method is suitable for an RS dataflow architecture, where partial sums are accumulated across multiple PEs, and hence, not suitable to perform gather for unique partial sums across the PEs, which exist in an OS dataflow architecture. Other research \cite{rethinking} proposes an array of microswitches that are configured to handle different kinds of DNN traffic by creating a tree. In ClosNN \cite{closnn}, one or more layers can be conducted on the network by mapping the neurons (PEs) on the input/output ports. Various stages of switching are done in order to connect the input and output ports in ClosNN, depending on the type of data traffic.
	\par
	Since the field of DNNs is evolving rapidly, hardware design should also be able to maintain this pace. As widely adopted NoC topology, mesh is used in most of the recently proposed DNN accelerators \cite{cerebras} \cite{groq} \cite{neuronlink} \cite{nocnn} \cite{mapping}. As many-to-one and one-to-many traffic are not inherently supported in a mesh topology, one-to-many has many solutions \cite{surveymulticast} that can be well adapted to DNN workloads, but many-to-one does not have an efficient method. Recent work modifies the topology to simulate a tree or Clos network \cite{closnn} \cite{rethinking} to support many-to-one traffic, while other work \cite{mapping}  \cite{yingwang} models this traffic as repetitive unicast (RU). In this work, we focus on providing communication support to this traffic on a mesh topology rather than proposing an alternate topology.  
	
	\section{Background and Motivation}\label{sec:motivation}
	
	Our study is focused on the inference phase of a feed-forward neural network. In general, the traffic patterns existing in a workload running on an NoC-based system significantly impact overall system performance \cite{treemulticast}. Hence, it is important to study the nature of traffic in a DNN workload and provide a communication mechanism to support this traffic efficiently. 
	
	\begin{figure}
		\centering
		\includegraphics[width=0.65\linewidth]{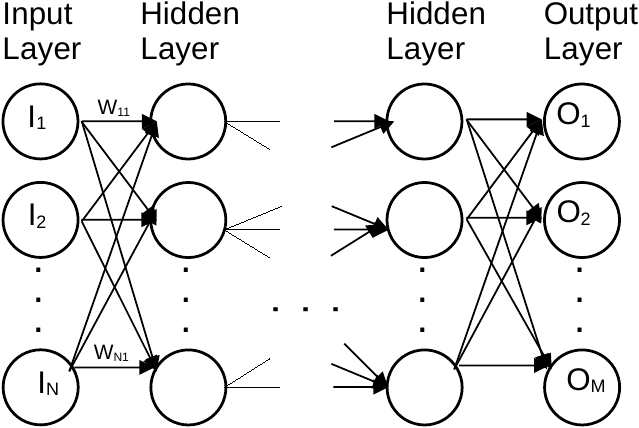}
		\caption{Example DNN model.}
		\label{fig-exampleDNN}
	\end{figure}	
	
\begin{figure}
		\centering
		\includegraphics[width=0.85\linewidth]{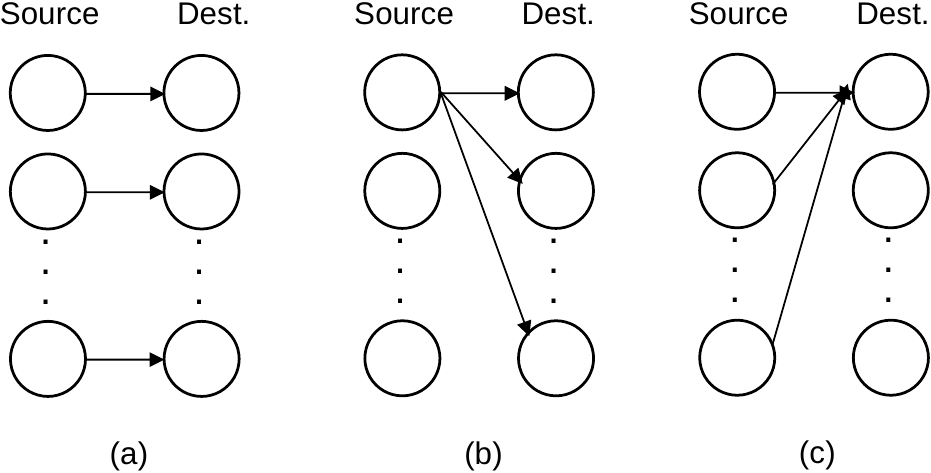}
		\caption{Traffic pattern inside a DNN hardware (a) Unicast (b) Multicast (c) Gather}
		\label{fig-traffic}
	\end{figure}

	A DNN model may include tens of layers (such as convolutional layers, pooling layers, and fully connected layers) and millions of parameters. The neurons (activations) in each layer are connected to neurons (activations) in another layer in full or in part via synapses (weights), as shown in Fig. \ref{fig-exampleDNN}. While implementing these layers in hardware, neurons are typically mapped to PEs inside a DNN accelerator. These neurons share the weights stored in the memory element; similarly, the outputs of the neurons in one layer is the input to the neurons in the adjacent layer. This sharing of data between adjacent PEs (neurons) creates traffic inside accelerators, which can be classified as one-to-one (unicast), one-to-many (multicast), and many-to-one (gather), as shown in Fig. \ref{fig-traffic}.
	
	\par
	Unicast traffic usually occurs when sending an input activation or weight from a memory element to a PE, or any other inter-PE traffic. Multicast traffic mainly covers the distribution of weights from the memory element to multiple PEs. Different dataflow mechanisms can be used to support multicast traffic for weight distributions. Gather traffic is used to collect the output from multiple PEs to the memory element. Due to limitations of computing resources, the inference operation of a DNN workload is performed in multiple rounds. When one round of MAC operations is completed, the intermediate results are gathered back to the memory element before initiating a new round.
	\par
	The output of each neuron in Fig. \ref{fig-exampleDNN} can be expressed as the operation shown in (\ref{eq1}).
	\begin{equation} \label{eq1}
	Output = \mathcal{F} \Bigg(\sum_{i=0}^{N-1}  I_{i} \cdot W_{i,1} + b \Bigg)
	\end{equation}
	where $W_{i,1}$ represents the weights and $I_i$ represents the input activation for the neurons in a particular layer containing $N$ neurons. $Output$ represents the output activation, which will be fed as input to another layer, and $\mathcal{F}()$ is the activation function in the model. Many DNNs consist of multiple layers, where both convolution and fully-connected layers perform MAC operations, as shown in (\ref{eq1}). These layers are computationally intensive, and hence, performed in multiple PEs in a distributed way. Moreover, the weights and inputs are stored in the memory element, and these steps require frequent access to the memory element, which is an expensive task in terms of latency and energy \cite{hardwareforML}. 
	\par
\begin{figure}
		\centering
		\includegraphics[width=0.95\linewidth]{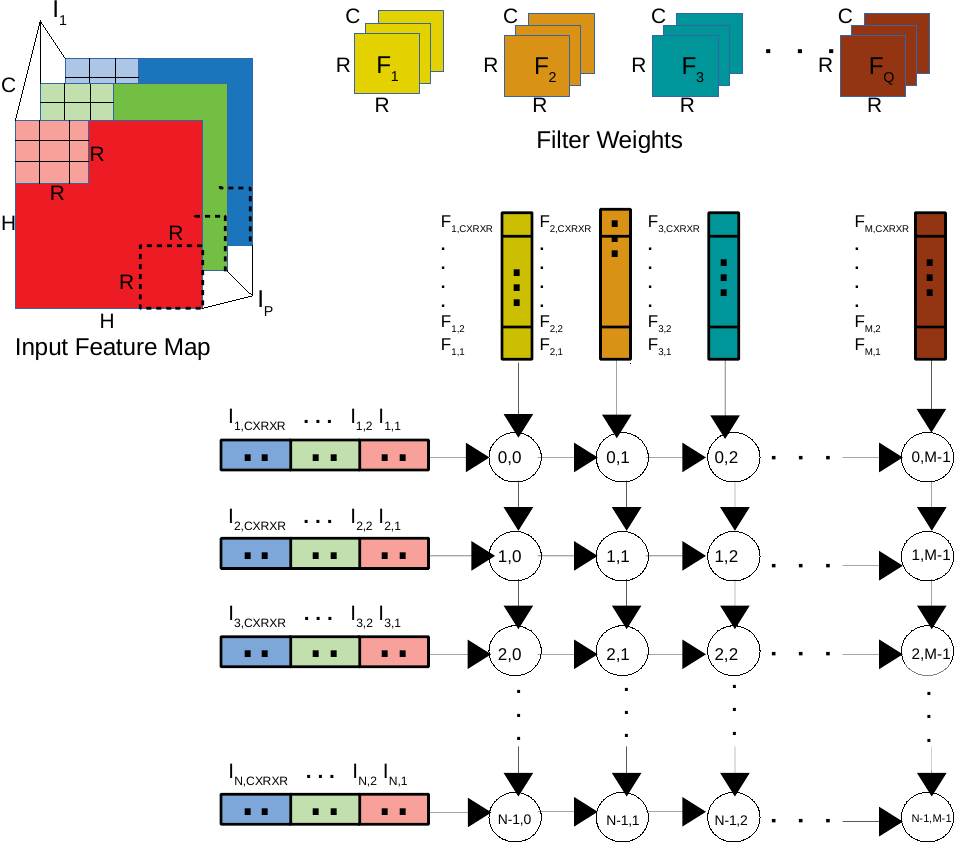}
		\caption{Dataflow in NxM Mesh NoC.}
		\label{fig-os}
	\end{figure}

	When designing a DNN hardware accelerator, dataflow is another crucial aspect to consider. It depends on many factors such as the input size, as well as the number of kernels, stride, and mapping scheme of the DNN workload onto the PE arrays, along with DNN optimizations like pruning and sparsity \cite{maeri}. An inefficient dataflow model will cause stalls, as appropriate data may not be available at the PE when needed, and low data reuse, so that the same data must be fetched multiple times from the memory, thus resulting higher latency and energy inefficiency. Compared with other dataflow models, the OS dataflow model achieves good performance with less complexity. In this work, we analyze our proposed method using the OS dataflow model on a mesh-based NoC. Fig. \ref{fig-os} shows the OS dataflow, where input activations and weights are streamed in a row-wise and column-wise manner, respectively, while the partial sums are accumulated on a PE. 
	\par
\begin{figure}
		\centering
		\includegraphics[width=0.9\linewidth]{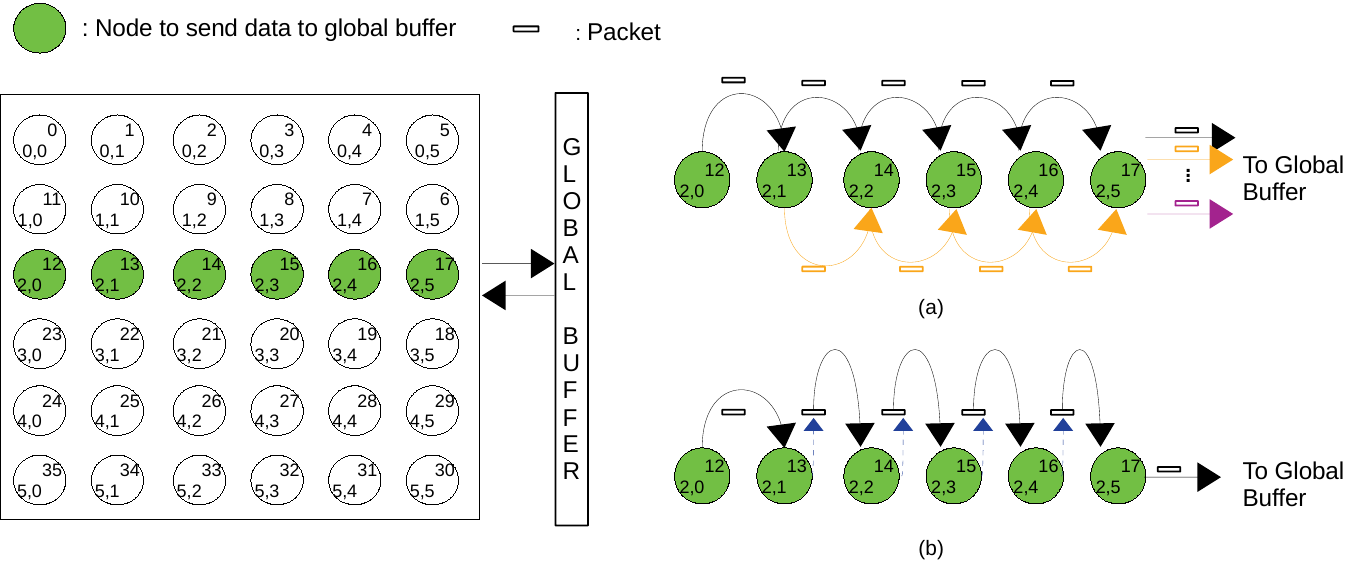}
		\caption{6x6 mesh example (a) without gather support (b) with gather support}
		\label{fig-motivation}
	\end{figure}

	Fig. \ref{fig-motivation} shows an example of how efficient communication support can affect the performance in a many-to-one type of communication traffic. The green-colored nodes in a 6x6 mesh are trying to send the data to a memory element. Fig. \ref{fig-motivation}(a) illustrates the delivery of data using unicast communication and the possible gather communication scheme, respectively. Using unicast communication, each node in the same row sends its packet to the same destination, which increases the amount of traffic. Is it possible to reduce the network traffic when multiple senders are sending a payload to the same destination? With the gather support, the gather packet is initiated at the left-most node and collects the data payload from the intermediate nodes on its way to the destination node. As shown in Fig. \ref{fig-motivation} (b), gather support significantly reduces the network traffic and reduces the total hop count from 15 to 5, proving to be efficient in delivering all data to the memory element using the least amount of resources. In addition, noticeably in DNN workloads, the weights and inputs used to calculate Eqn. (\ref{eq1}) are continuously streamed to certain groups of PEs, as shown in Fig. \ref{fig-os}. In that sense, direct links may be added for distributing weights/inputs, thus eliminating the unnecessary hop counts. As such, both many-to-one and one-to-many traffic can be supported in mesh-based NoC.
	
	\section{The Proposed Method}\label{sec:gather}
	In this section, we describe the gather supported routing scheme first, followed by the data streaming architectures and a support on multiple PEs per router. We also present the analysis of the performance improvement in this section.
	
	\subsection{Gather Support} \label{subsec:routing}
	Assume that a convolutional layer is implemented on an $NXM$ mesh-based NoC. Multiple PEs perform MAC operations in a distributed fashion as shown in Fig. \ref{fig-os} where $P$ input activations of size $H \cdot H$, each with $C$ channels, are streamed in X-dimension. Similarly, $Q$ filter, or weight streams of size $R \cdot R$, each with $C$ channels, are streamed in Y-dimension. Both the weight and input streams meet at all PEs in the respective dimension, to complete a total of $C \cdot R \cdot R$ MAC operations. As shown in Fig. \ref{fig-os}, input activations and weights are streamed from the left and top side of the PE array respectively, so that the partial sums are generated in PEs. It is also clear from Fig. \ref{fig-os} that multiple rounds are required to complete all MACs, due to the resource limitation in the network. When the first round of MAC operations are completed, the partial sum (PS) result, as shown in Eqn. (\ref{eq2}), is collected using gather traffic and sent to the global buffer before the start of the next round. Additionally, after the completion of one layer, output activations will also be moved to the memory element so that a new layer can operate on the PEs.
	
	\begin{equation} \label{eq2}
	\underset{i \in \{1...P\};k \in \{1...Q\}}{PS_{i,k}} = \sum_{j=1}^{C \cdot R \cdot R} (I_{i,j} \cdot F_{k,j} )
	\end{equation}
	
	\par
\begin{figure}
		\centering
		\includegraphics[width=1\linewidth]{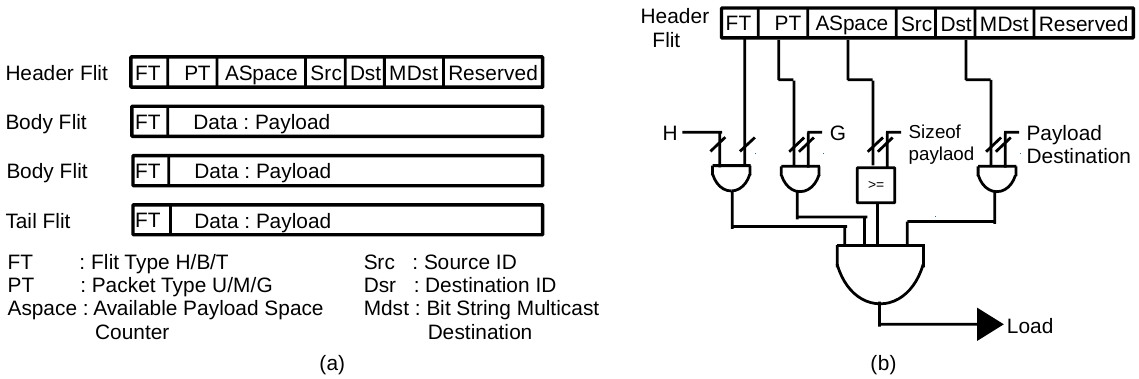}
		\caption{(a) Packet format (b) Load signal generation}
		\label{fig-packet}
	\end{figure}

	We propose to use gather-supported routing to support gather traffic. The leftmost PE in each row after completing the operation (PS or output activation) will initiate a gather packet, with the packet format as shown in Fig. \ref{fig-packet} (a). The packet consists of multiple fields including $FT$ to identify the flit type (head, body, or tail); $PT$ to identify the packet type (unicast, multicast, or gather);  $ASpace$ to indicate the available space for the gather payload;and $Src$ and $Dst$ to indicate the source and destination address and $MDst$ for multicast destinations, respectively. 
	\par
	Algorithm \ref{algorithm-flow} shows the flow for the gather supported routing implemented at each router. The incoming header flit is used to generate a $Load$ signal, which indicates the router to fill a gather payload in an incoming body or a tail flit by appending the payload. Ideally, the size of a gather payload is considered to be less than a flit size. Fig. \ref{fig-packet}(b) shows the logic to generate a $Load$ signal; the same signal is also used to decrement the space counter $ASpace$ so that other PEs can estimate the space for filling their gather payloads. If the $ASpace$ is less than a gather payload size, the router can initiate its own gather packet. However, to avoid the flooding of gather packets, each router must wait for the timeout period of $\delta$ cycle so that any other previously generated gather packet can go through. 
	\par
	The value of $\delta$ can be determined based on the router pipeline stages. Additionally, $\delta$ can be fine-tuned further for an individual router, if required. A too low value of $\delta$ will result in an increased amount of packets in the network, leading to congestion and increased latency, while a too high value of $\delta$ will cause nodes to wait too long for an incoming gather packet, which may increase the latency of the packets. Noticeably, $\delta$ also serves as a fault tolerance mechanism. If a link is faulty, then the node can initiate its own packet without having to wait indefinitely for a previously generated gather packet. In such a scenario, a large value of $\delta$ can lead to higher packet latency. In our experiments, all links are assumed to be fault free and reliable. 
	\par
	It is important to restrict a circular path in the routing algorithm to avoid a potential deadlock. The proposed gather packet still follows XY routing, which is deadlock-free.

	\begin{algorithm}
		\DontPrintSemicolon
		\SetAlgoLined
		\SetNoFillComment
		
		\SetKwInOut{Input}{Input}
		\SetKwInOut{Output}{Output}
		\Input{Arriving flit ($F$), Gather Payload ($P$) }
		\Output{Updated flit ($F$) or initiate a gather packet}

		\If{(($F.FT$ = $H$) and ($F.ASpace$ $>=$ $sizeof(P)$) and ($F.PT$ = $G$))}
		{	\tcp*[h]{generate a load signal}  
			
			\lIf{($F.Dst$ = $P.Dst$)} 
			{Load $\leftarrow$ 1} 
			\tcp*[h]{update F.ASpace before switch traversal}
			
			\lIf{(Load = 1)}{
				$F.ASpace \leftarrow$ $F.ASpace$ - $sizeof(P)$
			}
		}
		\lIf{(($F.FT$ = $B\ or\ F.FT$ = $T$) and (Load = 1))}
		{
			$F.Data \leftarrow$ $P.Data$
		}
		\lElse{$Can\ initiate\ a\ packet$}
		\lIf{($F.FT$ = $T$)}
		{
			Load $\leftarrow$ 0
		}
		\vspace{-0.1cm}
		\caption{Flow for the Gather Routing}
		\label{algorithm-flow}
	\end{algorithm}

	\subsection{Router Architecture} \label{subsec:arch}
\begin{figure}
		\centering
		\includegraphics[width=0.9\linewidth]{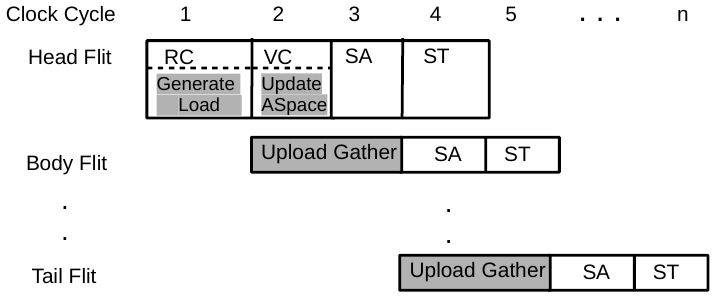}
		\caption{Modified router pipeline.}
		\label{fig-pipeline}
	\end{figure}

	An NoC router typically consists of multiple pipeline stages. Fig. \ref{fig-pipeline} shows a four-stage router pipeline including stages of route computation (RC), virtual channel allocation (VC), switch allocation (SA), and switch traversal (ST) \cite{book-dally}. For an incoming packet, only the header flit undergoes the RC stage to determine the output port for the packet. Similarly, only the header flit moves to the VC stage, where the flit arbitrates for a virtual channel corresponding to its output port. In the SA stage, each flit arbitrates for the switch input and output ports. Finally, in the ST stage, each flit traverses the crossbar. All flits of a packet undergo the SA and ST stages. The unused pipeline stages for the body/tail flit can be used to fill the gather payload into a gather packet.  
	\par
\begin{figure}
		\centering
		\includegraphics[width=0.9\linewidth]{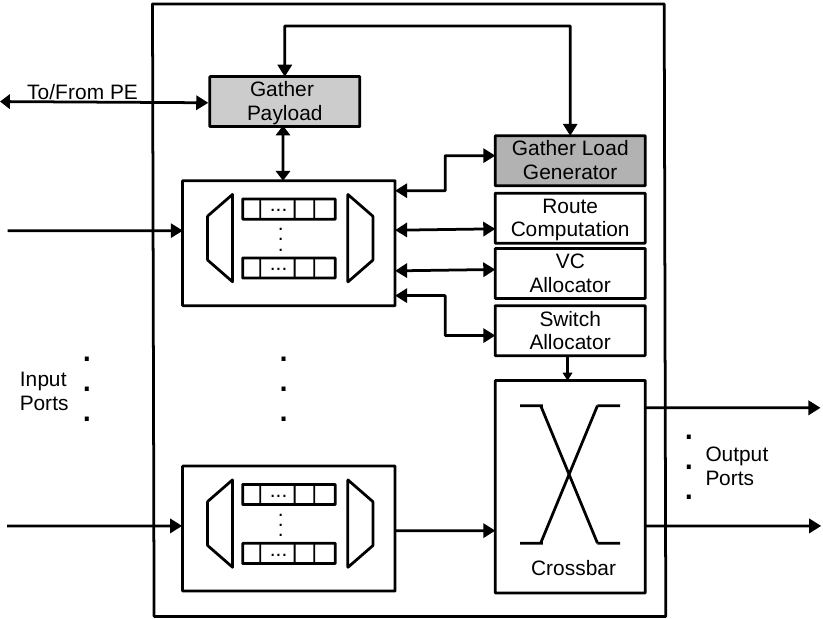}
		\caption{Router microarchitecture.}
		\label{fig-microarchitecure}
	\end{figure}

	Fig. \ref{fig-pipeline} shows the modified router pipeline to incorporate the gather support. When the header flit of a gather packet arrives at the input buffer, the $Load$ signal is generated during the RC stage and in the VC stage, and the $ASpace$ counter is updated. Upon the arrival of the body/tail flit, the gather payload is filled into the packet during the RC and VC stages. This modification does not require the packet to leave the router, nor add extra pipeline stages; thus, no additional latency is introduced. As the router pipeline is kept the same; there is no impact on the router performance. The modified router microarchitecture is shown in Fig. \ref{fig-microarchitecure}. The Gather Load Generator block will generate the $Load$ signal and update the $ASpace$ in the header flit. The Gather Payload block contains a queue that enqueues the gather payload from the PE, and the status of uploading is acknowledged back to the PE. The same status signal will be used by the PE to initiate its own gather packet if the incoming gather packet is full, or initiate its own unicast packet upon the expiration of $\delta$ cycles controlled by a counter set by the PE.

	\begin{figure}
		\centering
		\includegraphics[width=0.9\linewidth]{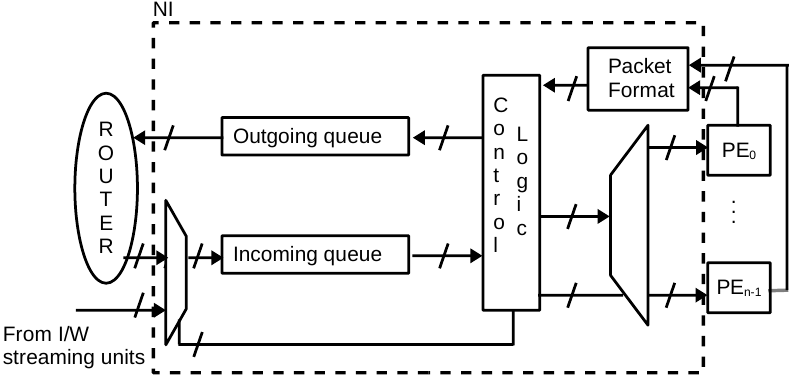}
		\caption{Multiple PE/router}
		\label{fig-PEs}
	\end{figure} 
	
	\subsection{Multicast Support} \label{subsec:modifieduarch}
	As shown in Fig. \ref{fig-os}, in the OS model, a partial sum will be accumulated at a PE with the input activations and weights streaming from the memory element \cite{shidiannao} \cite{os-moon}. Assume that one router can connect to up to $n$ PEs, so that these PEs will receive $n$ sets of inputs and weights. Convolution operation accounts for a large portion of energy and latency in a DNN operation \cite{hardwareforML}. A significant portion of the communication traffic involves the distribution of input activations and weights, which can be treated as multicast traffic because of the dataflow pattern, as shown in Fig. \ref{fig-os}. Based on this observation, a modification is proposed, where the input activations and weights are streamed using a bus from the memory elements to the PEs in the same row and column, respectively. The streaming bus will help in overcoming the additional routing overhead, and thus, improve the runtime latency of a DNN workload.
	\par
\begin{figure*}
		\centering
		\includegraphics[width=0.9\linewidth]{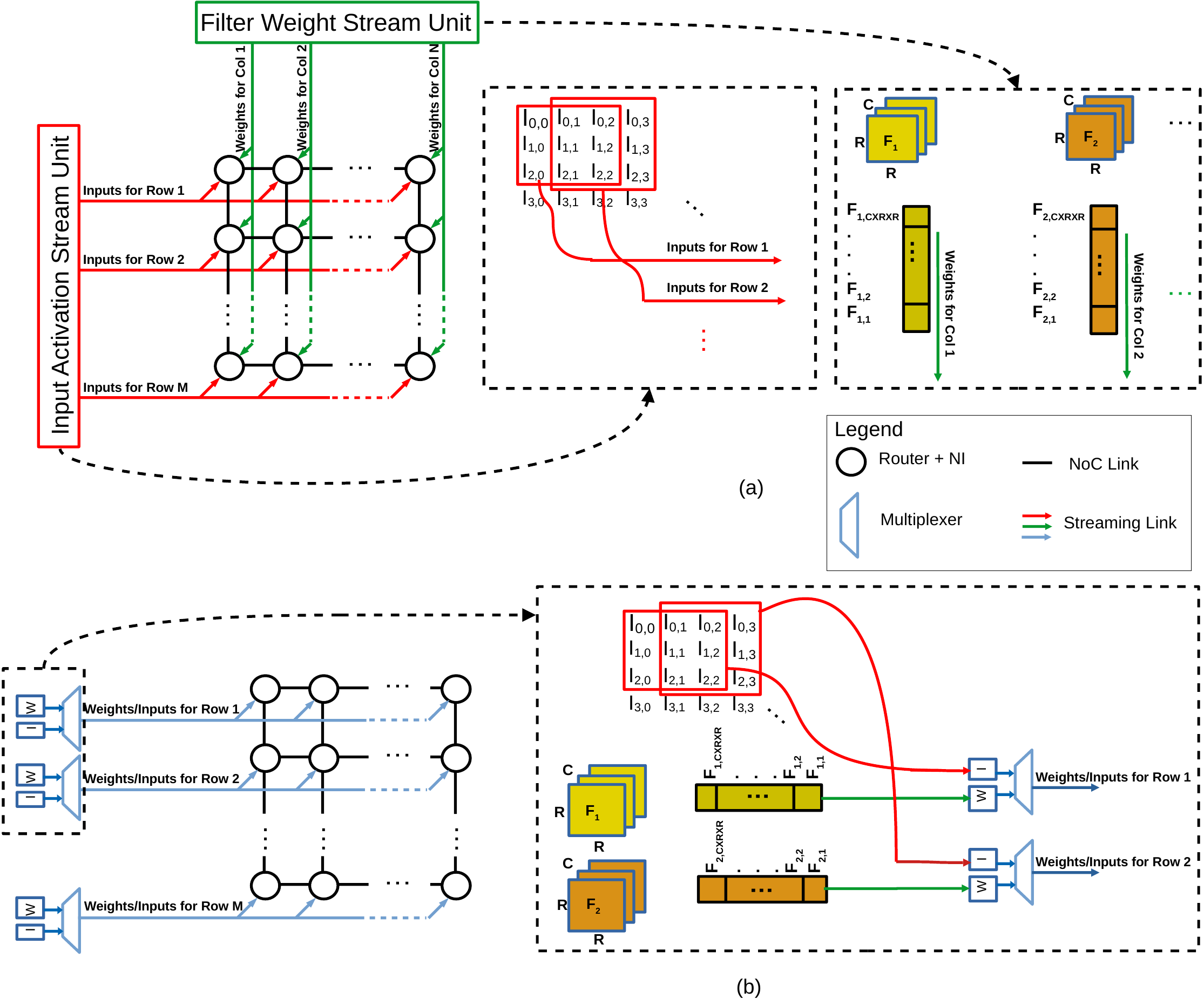}
		\caption{Modified architecture with direct streams (a) Two-way streaming (b) One-way streaming }
		\label{fig-modified_arch}
	\end{figure*}

	Fig. \ref{fig-modified_arch} (a) shows the two-way streaming architecture, where two different stream units will handle the streaming of input activations and filter weights. Each input activation streaming unit handles the streaming of the corresponding input activation from the memory element to a respective PE row. Each router in the same row will receive the same input activations, which are then buffered for MAC operation on a PE's internal register file. Similarly, the weight streaming unit streams the filter weights from the memory element to a respective PE column. All row and column data are streamed to the respective PEs, similar to the pattern shown in Fig. \ref{fig-os}. The partial sums or the output activations are calculated at all PEs. Results in the same row are then collected using a gather packet as it proceeds towards the memory element. Similar architectures can be orchestrated for other dataflow models.  
	\par
	Fig. \ref{fig-modified_arch} (b) shows the one-way streaming architecture, where both the input activations and the filter weights share the same streaming link to PEs in the same row. As the link is shared by either weight or the activation streams at a given clock cycle, there is inherent latency added before the PEs can move ahead with the MAC operation. Fig. \ref{fig-modified_arch}(b) also shows the streaming unit, which streams the input/weight activation in an interleaved manner through a multiplexor on the shared link. The partial sums will be accumulated by a gather packet before sending them back to the memory element. This architecture will use less silicon area compared with the two-way streaming architecture. This architecture may be beneficial for other types of dataflow models like WS, RS, where the weights are streamed first to the PEs before input streaming begins.     

	\begin{figure}
		\centering
		\includegraphics[width=0.7\linewidth]{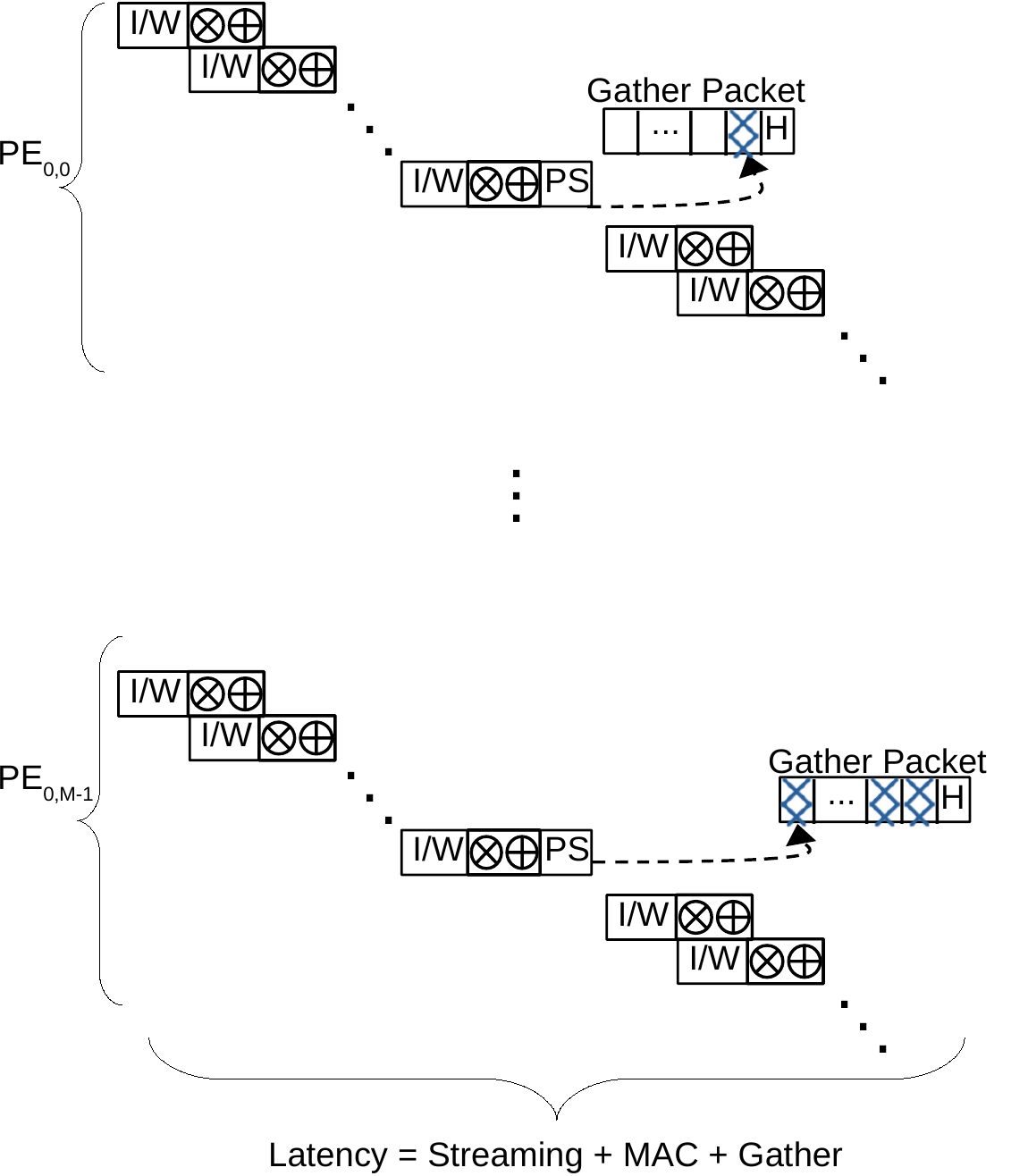}
		\caption{Pipelined operation of a partial sum(PS) generation/gather in a row of PEs}
		\label{fig-latency_chart}
	\end{figure}

	\subsection{Multiple PEs per Router} \label{subsec:networkInterface}
	We also consider an expanded mesh, where each router can be connected with multiple PEs. Fig. \ref{fig-PEs} shows such an architecture. The Network Interface (NI) unit handles the packet movement between the router and PEs. The streaming units feed input(s)/weight(s) (I/W) to the NI, as shown in Fig. \ref{fig-PEs}. An incoming packet from the router or streaming unit is dequeued from an incoming queue into a control logic, where the packet is disassembled and forwarded to the respective PEs. The control logic keeps track of the type of packet, start and end of the packet, and other necessary information to correctly decompose the data for a respective PE. When PEs are ready to inject data into the network, the packet format unit will collect the outgoing data from the PEs and generate a packet. The generated packet is then forwarded to the control logic, which creates flits to be enqueued in an outgoing queue. The router will access the outgoing queue and inject the packet into the network. Assume that these PEs are simple, as proposed in \cite{PE-model}, which supports MAC operation and an activation function with predictable pipeline stages. Hence, the synchronization can be done without extra overhead.
	\par
	Supporting multiple PEs per router allows more partial sums generated in parallel and makes better utilization of a gather payload, which can help accelerate the DNN execution with reduced power consumption. Depending on the bus width, multiple input activations and weights can be streamed in each NI at one time. As shown in Fig. \ref{fig-os}, these input activations and weights may have different combinations depending on how the PEs are grouped. One option is multiple PEs on the same column sharing one router; then multiple sets of input activation and one set of filter weight will be streamed in the NI. For example, for two PEs/router, $(I_{1,1}...I_{1,CXRXR})$, $(I_{2,1}...I_{2,CXRXR})$, and $(F_{1,1}...F_{1,CXRXR})$ will be streamed in the NI connected to $PE_{0,0}$ and $PE_{0,1}$ over multiple clock cycles. This can be further extended for 4 and 8 PEs/router. Another option is multiple PEs on the same row sharing one router; then one set of input activations and multiple sets of filter weights will be streamed in the NI. Other options are possible, with the cost of more complex design at the control logic. The information can be fed as a configuration file to the streaming units.
\par
In the proposed method, we have two different networks: one for gather traffic and the other for a streaming bus. In the mesh network, we use a credit-based flow control mechanism \cite{book-dally}. A similar end-end flow control mechanism may be employed for the streaming bus, but this may create an extra wire overhead from each node to the streaming unit. Hence, we adopt a similar credit-based mechanism to that used in \cite{rethinking} to ensure the single-cycle data delivery to the PEs. The global buffer maintains the status of the credits for the PEs, i.e, incoming queue in the NI, as shown in Fig. \ref{fig-PEs}. The streaming unit will only perform the streaming if all the nodes have free space to hold the data. This ensures the integrity of the MAC operation.  
	
	\subsection{Analysis of proposed modification} \label{subsec:modeling}
	The total clock cycles required to finish one round of convolution operation can be attributed to the time required for: input activation streaming and weight streaming; the MAC operation; and finally, the generation and collection of the result. Fig. \ref{fig-latency_chart} illustrates the pipelined execution of multiple rounds of convolution operations in one row of PEs, assuming one PE/router, where the streaming of input activation and weights (I/W), followed by the MAC operation and activation function, happen in parallel at all PEs. After $CXRXR$ MAC operations, the partial sum (PS) is generated at each PE, which is then collected by the gather packet. While the gather packet collects the PS results along its way to the global memory, the next round of convolution operation occurs concurrently. Note that with multiple PEs/router, in each round the streaming time will be extended while other parts stay the same.  
	\par
	Equations (\ref{eq3})-(\ref{eq4}) analyze the improvement of the runtime latency of a convolution layer using gather support over repetitive unicast (baseline) for the OS dataflow model on the proposed streaming architecture. In these equations, $C \cdot R \cdot R$ represents the time to stream the inputs to the PE, as shown in Fig. \ref{fig-os}; $n$ represents the multiplying factor, which depends on the number of PEs per router; $f_l$ represents the factor that reduces the input streaming with the streaming bus in the proposed method; $T_{MAC}$ represents the computation time for the MAC operation; $\kappa$ represents the number of pipeline stages at each router (with each stage occupying one cycle); and  $\frac{P}{N} \cdot \frac{Q}{M} \cdot \frac{1}{n}$ represents the number of rounds needed to finish the convolution of all $P$ inputs and $Q$ filters using the OS dataflow model. Assume that each unicast packet size is $L$, each gather packet size is $L'$, and the flit size is $W$. The gather packet is initiated from the leftmost node of each row in Fig. \ref{fig-os}. 
	
	\textit{$Latency_{_{RU}}$ =} 
	\begin{equation} \label{eq3}
	\resizebox{1\hsize}{!}{$
		\Big( \frac{C \cdot R \cdot R \cdot n}{f_l} + T_{MAC}\Big)\frac{P}{N} \cdot \frac{Q}{M} \cdot \frac{1}{n}  + M \cdot \kappa +  \Big \lceil \frac{L}{W} \Big \rceil  - 1 +  \Delta_R 
		$}
	\end{equation}
	Equation (\ref{eq3}) derives the runtime latency of a convolution layer using repetitive unicast, where $M \cdot \kappa$ represents the latency for the header flit of the result packet (partial sum) from $PE_{0,0}$ to reach the global buffer, $\Big \lceil \frac{L}{W} \Big \rceil -1$ represents the time needed for the remaining flits to arrive at the global memory, and $\Delta_R$ is the latency added due to the congestion. When a data streaming bus is used, as the transmission of unicast packets, all nodes are parallelized, the packet from the leftmost node will take the longest time to arrive at the global memory.
	\\
	
	\textit{$Latency_{_G}$ =} 
	\begin{equation} \label{eq4}
	\resizebox{1\hsize}{!}{$
		\Big(\frac{C \cdot R \cdot R \cdot n}{f_l}+ T_{MAC} \Big)\frac{P}{N} \cdot \frac{Q}{M} \cdot \frac{1}{n} +\sum_{i = 0}^{\big \lceil \frac{M.n}{\eta}\big \rceil - 1}  \big(( M - i \cdot \frac{\eta}{n}) \cdot \kappa + \big \lceil \frac{L'}{W} \big \rceil -1 \big) +  \Delta_G $}
	\end{equation}
	Equation (\ref{eq4}) derives the runtime latency of a convolution layer using gather support, where $\eta$ is the number of payloads that can be collected by one gather packet, $\big \lceil \frac{M.n}{\eta}\big \rceil $ represents the number of gather packets, $(M - i \cdot \frac{\eta}{n}) \cdot \kappa$ represents the latency for the header flit in the gather packet, $\big \lceil\frac{L'}{W} \big \rceil-1$ represents the latency for the rest flits in the gather packet, and $\Delta_G$ is the latency added due to the congestion.
	
\par
Note that in Eqs. \ref{eq3} and \ref{eq4}, the data streaming and MAC operation time is same; the difference lies in the time taken to transmit the results to the global memory. When n=1, the time taken to transmit the unicast packet from the leftmost node is nearly the same as the time taken to transmit the gather packet. However, when n increases, the delay due to network congestion will increase significantly for RU (reflected by $\Delta_R$). In comparison, the network congestion for gather packets (reflected by $\Delta_G$) will be much less. We will evaluate the effects of $\Delta_R$ and $\Delta_G$ through simulations in Section \ref{sec:evaluationdelta}.   

	\section{Performance Evaluation}\label{sec:evaluation}
	To evaluate the performance of the proposed method, simulations are conducted and compared with the repetitive unicast method on mesh-based NoCs modified with the streaming architectures for the OS dataflow model. In this section, the experiment settings are described first, followed by the study of timeout period ($\delta$) and gather packet size, before the performance analysis is presented.  
	
	\subsection{Experiment Setup}
	We compare the proposed method and repetitive unicast method in terms of the runtime latency and power consumption. We assume that there is a higher level entity or a mapping framework similar to  \cite{tpu} \cite{shidiannao} \cite{neuronlink} \cite{closnn} \cite{yingwang} exists that does the task of mapping neurons to the PEs, controlling timing for better synchronization without stalls, so that our focus is on evaluating the performance of the on-chip network. In order to fully utilize the spatial PE arrays, the parameters obtained from Pytorch framework \cite{pytorch} are used to model the traces for the NoC. PEs represent the neurons that are organized in a 2D mesh. The total neurons in each layer are divided to fit the PEs in a mesh. Assume that memory elements (global memory) are located on the north, east, and west sides of the network. Each PE receives the input activation and filter weights from the streaming units on the north and east sides of the mesh, respectively. Accumulation happens locally to generate the partial sums, which are then collected from the left side to the global buffer at the right side of the mesh. The output feature map of the current layer is completely generated before moving ahead with another layer. 
	\begin{table}[]
		\caption{Network Configuration}
		\begin{center}
			\begin{tabular}{|l|l|}
				\hline 
				Topology         & 8x8 Mesh, 16x16 Mesh  
				\\ \hline
				Virtual Channels & 2                                                                                       
				\\ \hline
				Latency     & router: 4 cycles, link: 1 cycle                 
				\\ \hline
				Buffer Depth     & 4 flits   
				\\ \hline
				Flit Size        & 128 bits/flit                                                                            
				\\ \hline
				Gather Payload   & 32 bits                                                                                  
				\\ \hline
				Number of PE per router	& 1,2,4,8	
				\\ \hline
				Gather Packet Size      & \begin{tabular}[c]{@{}l@{}}3,5,9,17 flits/packet for\\ 1,2,4,8 PEs/router resp.\end{tabular} 
				\\ \hline
				Unicast Packet Size      & \begin{tabular}[c]{@{}l@{}}2 flits/packet \end{tabular}
                 \\ \hline
				$T_{MAC}$	& 5	
				\\ \hline
				
			\end{tabular}
		\end{center}
		\label{table-network}
	\end{table}
	
	\par 
	A cycle-accurate C++ based NoC simulator \cite{popnet} is used to simulate the generated traces for Alexnet \cite{alexnet} and VGG-16 \cite{vgg16}. Orion 3.0 \cite{orion} is used to estimate the power consumption for NoC, and DSENT \cite{dsent} is used to estimate the power consumption for the streaming bus. Simulations are performed on 8x8 and 16x16 2D mesh networks. Table \ref{table-network} shows the NoC setting used for performance analysis. As the number of PEs/router increases, the gather payload also increases. To accommodate these gather payloads, we can either use a fixed number of flits or a dynamic flit size per gather packet. For the DNN workload, each node in the same row will generate a prefix sum result. Hence, a fixed number of flits is chosen for our experiments, which avoids the extra overhead in router design if a dynamic flit size is used. For 1,2,4,8 PEs/router, the gather packet size is set as 3,5,9,17 flits/packet by default, respectively. This flit size is enough to collect all the gather payloads for an 8x8 network; however, for a 16x16 NoC, two gather packets are needed, as the first one will be full halfway to the global memory.

	\begin{figure}
		\centering
		\includegraphics[width=1\linewidth]{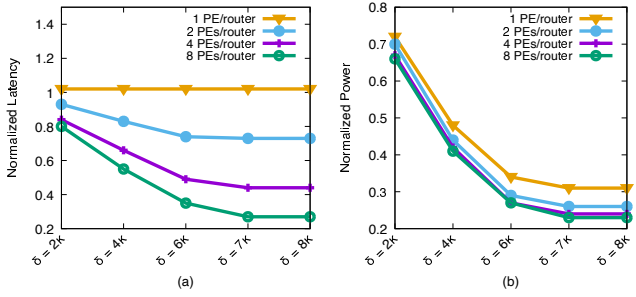}
		\caption{Analysis of $\delta$ on 8x8 mesh for different number of PEs/router}
		\label{fig-delta}
	\end{figure}

	\begin{figure*}
		\centering
		\includegraphics[width=1\linewidth]{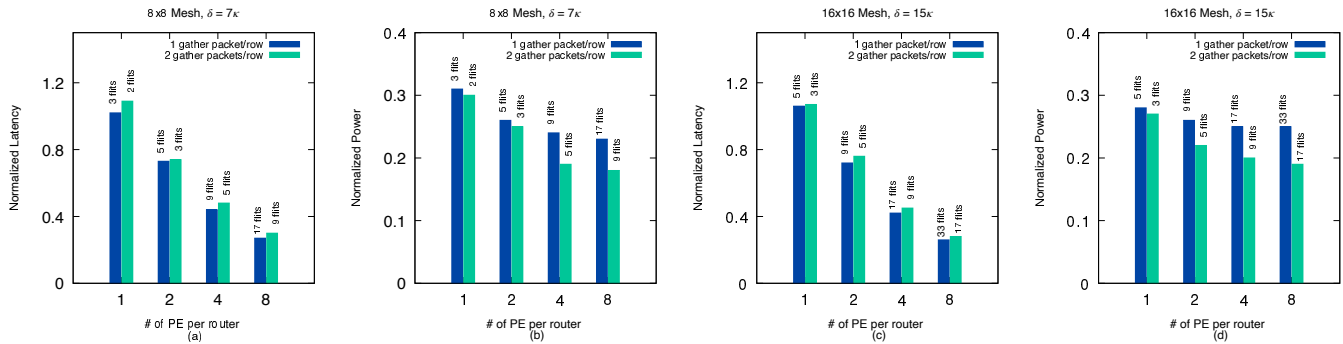}
		\caption{Analysis of different gather packet size on 8x8 mesh (a),(b) and 16x16 mesh (c),(d) for different number of PEs/router}
		\label{fig-packetAnalysis}
	\end{figure*}
	
	\subsection{Analysis of $\delta$ and Gather Packet Size} \label{sec:evaluationdelta}
	The timeout period $\delta$ plays an important role in the performance of the gather routing. The $\delta$ value defines the waiting time (in cycles) for a PE with a gather payload to wait before initiating its own packet, with the anticipation that the gather packet sent from its neighbor will arrive. Fig. \ref{fig-delta} shows the impact of $\delta$ on the total runtime latency, as well as the total power consumption of gather supported routing. The analysis is done on an 8x8 mesh under a similar traffic scenario as in Fig. \ref{fig-motivation}, where the nodes in one row are trying to deliver the gather payload to the global memory on the right side of the mesh. 
	\par
	The time out period ($\delta$ clock cycles) actually depends on the router pipeline stages ($\kappa$). When $\delta<\kappa$, the header flit of a gather packet will not reach its adjacent node before the expiration of the $\delta$ clock cycle. As such, each PE will initiate its own packet. This situation is similar to sending the result from each node in a repetitive unicast way, which will cause congestion in the network and result in increased runtime latency. As the value of $\delta$ increases, the gather packet initiated by a PE will subsequently reach the adjacent nodes before timeout occurs at those nodes. This piggyback mechanism will effectively decrease the network traffic and improve the resource utilization in the NoC. 
	\par
	As shown in Fig. \ref{fig-delta}(a), the normalized runtime latency (vs. $\delta<\kappa$) is reduced with $\delta$ value increased except for the case with 1PE/router, which is almost same. With more PEs/router, the gather packer size is increased (3, 5, 9, 17 flits for 1, 2, 4, 8 PEs/router) to accommodate more partial sums. Noticeably, after $\delta$ becomes sufficiently large ($7\kappa$), further latency improvement is not noticed. This is because the $\delta$ value is large enough to allow all the gather payload to be collected by a gather packet. Therefore, for $NXN$ mesh, $\delta$ can be set to $(N-1)\kappa$ to ensure that the header flit of the leftmost gather packet will arrive at all nodes in the same row, so that all the gather payloads can be uploaded into the same gather packet. 
	\par
	Fig. \ref{fig-delta}(b) shows the normalized power consumption for different values of $\delta$ (vs. $\delta<\kappa$). With increasing the $\delta$ value, the number of packets generated in the network is reduced as the gather packet is collecting all the partial sum results, which helps reduce the total number of hops traversed and optimizes the NoC resource utilization for 1,2,4,8 PEs/router, thus consuming less total power. Although, there is not any improvement in the runtime latency for the 1PE/router case, we see some significant improvement in the power consumption.  
	\par
	The tradeoff of using different gather packet sizes is further studied for different network sizes as well as numbers of PEs/router settings. Fig. \ref{fig-packetAnalysis} compares the performance of gather traffic using one gather packet with a larger number of flits and two gather packets with smaller number of flits on 8x8 and 16x16 mesh for different numbers of PEs/router. Fig. \ref{fig-packetAnalysis}(a) and (b) shows the normalized runtime packet latency and power consumption (vs. reptitive unicast) for 8x8 mesh. We can see a clear tradeoff, where using one gather packet with larger number of flits is better in terms of latency improvement, but using two gather packets with a smaller number of flits is better at improving the power consumption. A similar trend on 16x16 mesh is also shown in Fig. \ref{fig-packetAnalysis}(c) and (d). 
	\par
	For the 1 PE/router case, a slightly increase in runtime latency occurs, as this is the in which the network does not have much load for the $RU$ case, i.e, one packet per node. In addition, we notice that with smaller gather flit size, one can expect small runtime; however, it is the opposite, as shown in Figs. \ref{fig-packetAnalysis}(a) and (b). This effect is seen due to the expiration of $\delta$ clock cycles; for the 2 gather packets per row case, the second packet is only injected when the first packet reaches the node, with no space left for further payload. In such a case, the first node to encounter such a situation will initiate a new gather packet, which is the second gather packet. To avoid this scenario, the router can be hardwired with the information to initiate the gather packet without waiting for the incoming gather packet, which reduces the scope and scalibility of the proposed method. To balance the tradeoff of latency and power consumption, in the following performance analysis, one gather packet is used for 8x8 mesh and 2 gather packets are used for 16x16 mesh with 3, 5, 9, 17 flits/gather packet set for 1, 2, 4, 8 PEs/router, respectively.  
	
	\begin{figure}
		\centering
		\includegraphics[width=0.9\linewidth]{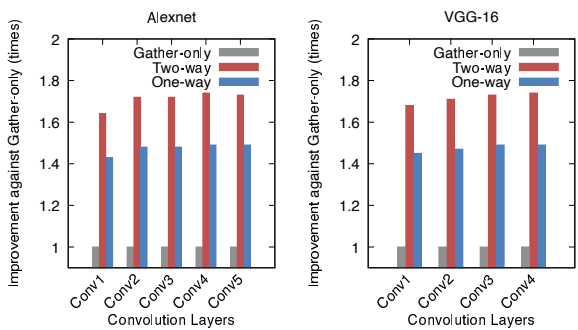}
		\caption{Simulated improvement on the runtime latency of different convolution layers in Alexnet \cite{alexnet} and VGG-16 \cite{vgg16} over Gather-only \cite{socc}}
		\label{fig-improvement_simulated}
	\end{figure}   
	
	\begin{figure*}
		\centering
		\includegraphics[width=1\linewidth]{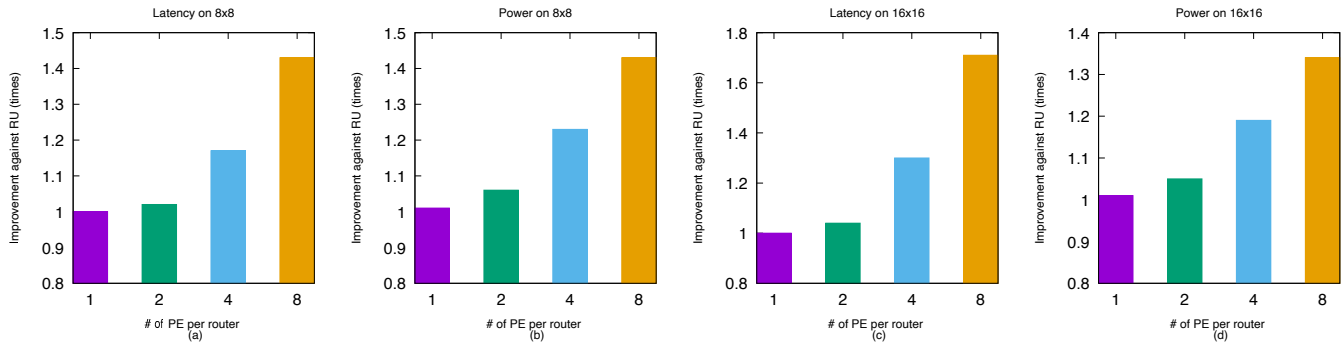}
		\caption{Improvement on total runtime latency (a),(c) and power consumption (b),(d) for Alexnet \cite{alexnet} over RU for different number of PEs/router}
		\label{fig-resultAlexnet}
	\end{figure*}

	\begin{figure*}
		\centering
		\includegraphics[width=1\linewidth]{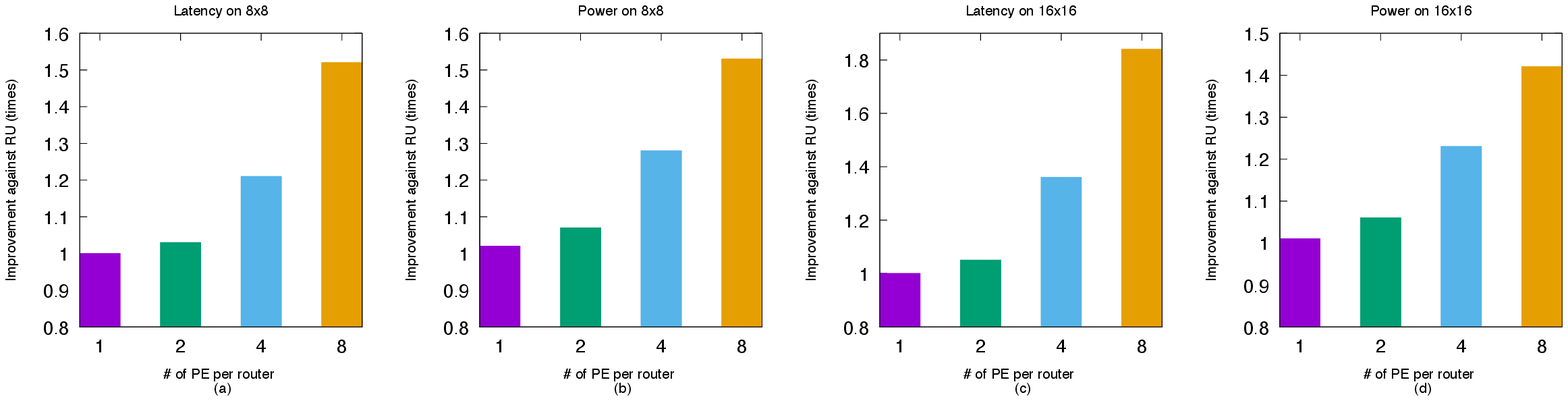}
		\caption{Improvement on total runtime latency (a),(c) and power consumption (b),(d) for VGG-16 \cite{vgg16} over RU for different number of PEs/router}
		\label{fig-resultVgg16}
	\end{figure*}

	\subsection{Performance Analysis}
	\par
	Fig. \ref{fig-improvement_simulated} shows the simulated runtime latency improvement of the proposed gather support with two-way streaming and one-way streaming architectures vs. gather-only using the NoC parameter from Table \ref{table-network}. On average, the gather support with two-way streaming architecture achieves 1.71 times improvement, and the gather support with one-way streaming obtains 1.48 times improvement compared against the gather-only method \cite{socc}. It is clear that the runtime latency improvement using two-way streaming is better than using one-way streaming in the OS dataflow model. Hence, we compare gather support against repetitive unicast with two-way streaming architecture in our experiments further.
\par
Figs. \ref{fig-resultAlexnet}(a) and (c) show the improvement in the total runtime latency of the proposed method against the repetitive unicast method for all convolution layers in Alexnet \cite{alexnet} on 8x8 and 16x16 mesh-based NoCs, respectively. It is clear that as the number of PEs is increased across 8x8 or 16x16 mesh, we can see an improvement in the total runtime latency. This improvement is attributed to more parallel operations enabled by an increasing number of PEs per router. With more PEs, more MAC operations are done in parallel in one round, which reduces the number of rounds needed. For a lower number of PEs per router, the runtime improvement is minor, which can also be seen from the delta analysis (Fig. \ref{fig-delta}(a)), as the network is not congested enough for the gather packet to improve the latency.
\par
The performance improvement is higher in the case of 16x16 mesh when compared with the 8x8 mesh. The reason is that on the 16x16 mesh, repetitive unicast traffic creates much higher congestion in the network, and the benefit of using gather traffic is more significant than on the 8x8 mesh. Similarly, Fig. \ref{fig-resultVgg16}(a) and (c) show the improvement in total runtime latency for all convolution layers in VGG-16 \cite{vgg16} for 8x8 and 16x16 meshes. For VGG-16, we see a similar trend in performance improvement, with the 16x16 mesh offering more improvement (up to 1.84 times) than the 8x8 mesh, and the improvement is better with the increasing number of PEs per router. On average, performance improvement is higher in VGG-16 compared with Alexnet, as it has a lot more convolution layers to process than Alexnet.
\par
Figs. \ref{fig-resultAlexnet}(b) and (d) shows the improvement in the total network power consumption of the proposed method against the repetitive unicast method for Alexnet on the 8x8 and 16x16 mesh-based NoCs. Different from runtime latency, network power consumption is determined by the total amount of traffic communicated. For a lower number of PEs per router, the power improvement is minor because the power consumption due to the streaming is higher than the power saving from the gather traffic. As the number of PEs/router increases, improvement in the total network power also increases (up to 1.4 times) because of the reduction in streaming power. More weights or inputs can be streamed with an increasing number of PEs/router, and the advantage of gather traffic over the repetitive unicast traffic is more significant. For the 16x16 mesh, we can see that the improvement is slightly less than the 8x8 mesh, which is due to the increased number of gather packets for the same of PEs/router. Similarly, Figs. \ref{fig-resultVgg16}(b) and (d) show improvement in power performance for VGG-16 \cite{vgg16} for 8x8 and 16x16 meshes. For both mesh sizes, we see a similar performance trend as in Fig. \ref{fig-resultAlexnet}(b) and (d). Similarly, the improvement of power consumption is higher in VGG-16 compared to Alexnet for the same reason as the latency improvement.

\subsection{Hardware Overhead}
We used DSENT \cite{dsent} to estimate the area and power for a router with the configuration shown in Table \ref{table-network}. For a router operating on a $1\ GHz$ clock, it consumes $26.3\ mW$ power with an area of $72106\ \mu m^2$. The hardware overhead of a proposed router from Fig. \ref{fig-microarchitecure} is evaluated with the synthesis report obtained from the Synopsys Design Compiler with a $45\ nm$ CMOS library. The power consumption of a proposed router is $27.87\ mW$ and the area is $74950 \ \mu m^2$. With the proposed changes in the router, the overhead is around a $6\%$ increase in power and $4\%$ increase in an area, which is worthwhile considering the performance improvement with the changes.    
	
	\section{Conclusion}\label{sec:conclusion}	
	In this paper, we proposed using the gather packet and direct data streaming architectures on mesh-based NoC to handle abundant many-to-one and one-to-many traffic in DNN workloads. The OS dataflow model is adopted to study the proposed method, which is evaluated using two DNN models: Alexnet and VGG-16. The analysis shows that the two-way streaming architecture achieves more significant improvement in the runtime latency of a convolutional layer. Simulation results confirm the effectiveness of the proposed method, which achieves up to 1.8 times improvement in the runtime latency and up to 1.7 times improvement in the network power consumption. The hardware overhead of the proposed method is justifiable for the performance improvements achieved over the repetitive unicast method. Future work should include applying the proposed method to other dataflow models, as well as thorough performance study of different dataflow models.

	\ifCLASSOPTIONcaptionsoff
	\newpage
	\fi

	
	

\begin{thebibliography}{1}
	
	\bibitem{deeplearning-nature}
	Y. LeCun, Y. Bengio, \& G. Hinton, "Deep learning," Nature, 2015, vol. 521,pp. 436–444. https://doi.org/10.1038/nature14539
	
	\bibitem{selfdriving}
	C. Chen, A. Seff, A. Kornhauser and J. Xiao, "DeepDriving: learning affordance for direct perception in autonomous driving," Proc. IEEE Int'l Conf. on Computer Vision (ICCV), Santiago, 2015, pp. 2722-2730.
	
	\bibitem{cancerdetection}
	Esteva, A., Kuprel, B., Novoa, R. et al., "Dermatologist-level classification of skin cancer with deep neural networks," Nature, 2017, vol. 542, pp. 115–118. https://doi.org/10.1038/nature21056
	
	\bibitem{alexnet}
	A. Krizhevsky, “One weird trick for parallelizing convolutional neural networks,” CoRR, vol. abs/1404.5997, 2014. [Online]. Available: http://arxiv.org/abs/1404.5997
	
	\bibitem{vgg16}
	K. Simonyan and A. Zisserman, “Very deep convolutional networks for large-scale image recognition,” in Proc. ICLR, 2015.
	
	\bibitem{hardwareforML}
	V. Sze, Y. Chen, J. Emer, A. Suleiman and Z. Zhang, "Hardware for machine learning: challenges and opportunities," in Proc. IEEE Custom Integrated Circuits Conf. (CICC), Austin, TX, 2017, pp. 1-8.
	
	\bibitem{noc-routepacketnotchips}
	W. J. Dally and B. Towles, “Route packets, not wires: on-chip interconnection networks,” in Proc. 38th DAC, 2001, pp. 684–689.
	
	\bibitem{maeri}
	Hyoukjun Kwon, Ananda Samajdar, and Tushar Krishna, "MAERI: enabling flexible dataflow mapping over DNN accelerators via reconfigurable interconnects", in Proc. ASPLOS, 2018, pp. 461–475.
	
	\bibitem{spikingNN}
	S. Carrillo et al., "Scalable hierarchical network-on-chip architecture for spiking neural network hardware implementations," IEEE Trans. Parallel and Distributed Systems, 2013, vol. 24, pp. 2451-2461.
	
	\bibitem{kilocore}
	B. Bohnenstiehl et al., "KiloCore: a 32-nm 1000-processor computational array," IEEE J. of Solid-State Circuits, 2017, vol. 52, no. 4, pp. 891-902.
	
	\bibitem{all-to-all}
	A. Touzene, "On all-to-all broadcast in dense gaussian network-on-chip," IEEE Trans. on Parallel and Distributed Systems,2015,  vol. 26, no. 4, pp. 1085-1095.
	
	\bibitem{pdp2020}
	B. Tiwari, M. Yang, Y. Jiang and X. Wang, "Efficient on-chip multicast routing based on dynamic partition merging," in Proc. 28th Euromicro Int'l Conf. on Parallel, Distributed and Network-Based Processing (PDP), Vasteras, Sweden, 2020, pp. 274-281.
	
	\bibitem{cerebras}
	Cerebras Systems, "Wafer-scale deep learning," in Proc. IEEE Hot Chips 31 Symp. (HCS), Cupertino, CA, 2019, pp. 1-31.
	
	\bibitem{groq}
	D. Abts et al., "Think Fast: a Tensor Streaming Processor (TSP) for accelerating deep learning workloads," in Proc. ACM/IEEE 47th Annual Int'l Symp. on Comp. Architecture (ISCA), Valencia, Spain, 2020, pp. 145-158.
	
	\bibitem{surveymulticast}
	A. Karkar, T. Mak, K. Tong and A. Yakovlev, "A survey of emerging interconnects for on-chip efficient multicast and broadcast in many-cores," IEEE Circuits and Systems Magazine, 2016, vol. 16, no. 1, pp. 58-72.
	
	\bibitem{tpu}
	Jouppi, N. P. et al., "In-datacenter performance analysis of a tensor processing unit", in Proc. 44th Int. Symp. Comp. Architecture (ISCA), 2017.
	
	\bibitem{eyeriss}
	Y. Chen, T. Yang, J. Emer and V. Sze, "Eyeriss v2: a flexible accelerator for emerging deep neural networks on mobile devices," IEEE J. on Emerging and Selected Topics in Circuits and Systems, 2019, vol. 9, no. 2, pp. 292-308.
	
	\bibitem{shidiannao}
	Z. Du et al., "ShiDianNao: shifting vision processing closer to the sensor," in Proc. ACM/IEEE 42nd Annual Int'l Symp. on Comp. Architecture (ISCA), Portland, OR, 2015, pp. 92-104.
	
	\bibitem{fpgadnn}
	H. Sharma et al., "From high-level deep neural models to FPGAs," in Proc. 49th Annual IEEE/ACM Int'l Symp. on Microarchitecture (MICRO), Taipei, 2016.
	
	\bibitem{binarizedNN}
	E. Nurvitadhi, D. Sheffield, Jaewoong Sim, A. Mishra, G. Venkatesh and D. Marr, "Accelerating binarized neural networks: comparison of FPGA, CPU, GPU, and ASIC," in Proc. Int'l Conf. on Field-Programmable Technology (FPT), Xi'an, 2016.
	
	\bibitem{diannao} 
	T. Chen et al., “DianNao: a small-footprint high-throughput accelerator for ubiquitous machine-learning,” in Proc. ASPLOS, 2014.
	
	\bibitem{dadiannao}
	T. Luo et al., "DaDianNao: a neural network supercomputer," IEEE Trans. on Computers, 2017, vol. 66, no. 1, pp. 73-88.
	
	\bibitem{nocnn}
	D. Vainbrand and R. Ginosar, "Network-on-chip architectures for neural networks,"in Proc. 4th Int'l Symp. Networks-on-Chip (NoCS), Grenoble, 2010, pp. 135-144.
	
	\bibitem{spinnaker}
	E. Painkras, et al., “SpiNNaker: a 1-W 18-core system-on-chip for massively-parallel neural network simulation,” IEEE J. Solid-State Circuits, 2013, vol. 48, no. 8, pp. 1943–1953.
	
	\bibitem{neuronlink}
	S. Xiao et al., "NeuronLink: an efficient chip-to-chip interconnect for large-scale neural network accelerators," IEEE Trans. on Very Large Scale Integration (VLSI) Systems, 2020, vol. 28, no. 9, pp. 1966-1978.
	
	\bibitem{closnn}
	R. Hojabr, M. Modarressi, M. Daneshtalab, A. Yasoubi and A. Khonsari, "Customizing Clos network-on-chip for neural networks," IEEE Trans. on Computers, 2017, vol. 66, no. 11, pp. 1865-1877.
	
	\bibitem{socc}
	B. Tiwari et al., "Improving the performance of a NoC-based CNN accelerator with gather support", in Proc. IEEE 33rd Intl. System-on-Chip Conf. (SOCC), 2020.
	
	\bibitem{neu-noc}
	X. Liu and et al., “Neu-NoC: a high-efficient interconnection network for accelerated neuromorphic systems,” in Proc. 23rd ASP-DAC, 2018, pp. 141–146.
	
	\bibitem{3dNN-acc}
	A. Firuzan, M. Modarressi, M. Daneshtalab, and M. Reshadi, “Reconfigurable network-on-chip for 3D neural network accelerators,” in Proc. 12th Int'l Symp. Networks-on-Chip (NoCS), 2018, pp. 1–8.
	
	\bibitem{rethinking}
	H. Kwon, A. Samajdar, and T. Krishna, “Rethinking NoCs for spatial neural network accelerators,” in Proc. 11th Int'l Symp. Networks-on-Chip (NoCS), 2017.
	
	\bibitem{mapping}
	Seyedeh Mirmahaleh et al.,"Flow mapping and data distribution on mesh-based deep learning accelerator," in Proc. 13th Int'l Symp. Networks-on-Chip (NoCS), 2019.
	
	\bibitem{yingwang}
	Ying Wang et al., "A many-core accelerator design for on-chip deep reinforcement learning," in Proc. 39th ICCAD, 2020, pp. 1-7.
	
	\bibitem{treemulticast}
	N. E. Jerger, L. Peh and M. Lipasti, "Virtual circuit tree multicasting: a case for on-chip hardware multicast support," in Proc. Int'l Symp. on Comp. Architecture (ISCA), Beijing, 2008, pp. 229-240.
	
	\bibitem{book-dally}
	W. Dally and B. Towles, "Principles and practices of interconnection networks", Morgan Kaufmann Publishers Inc., San
	Francisco, CA, USA, 2003.
	
	\bibitem{os-moon}
	B. Moons and M. Verhelst, “A 0.3–2.6 TOPS/W precision-scalable processor for real-time large-scale ConvNets,” in Proc. Symp. VLSI, 2016, pp. 1–2.
	
	\bibitem{PE-model}
	H. Esmaeilzadeh, A. Sampson, L. Ceze and D. Burger, "Neural acceleration for general-purpose approximate programs," in Proc. 45th Annual IEEE/ACM Int'l Symp. on Microarchitecture (MICRO), Vancouver, BC, Canada, 2012, pp. 449-460.
	
	\bibitem{pytorch}
	A. Paszke and et al., “PyTorch: An imperative style, high-performance deep learning library,” 33rd Conf. on Neural Information Processing Systems (NeurIPS), Vancouver, Canada, 2019, pp. 8024-8035.
	
	\bibitem{popnet}
	X. Wang, T. Mak, M.Yang, Y. Jiang, and et al., “On self-tuning networks-on-chip for dynamic network-flow dominance adaptation,” in Proc. 7th Int'l Symp. Networks-on-Chip (NoCS), 2013, pp. 1–8.
	
	\bibitem{orion}
	A. B. Kahng, B. Lin, and S. Nath, “ORION3.0: a comprehensive NoC router estimation tool,” IEEE Embedded Systems Letters, 2015, vol. 7, no. 2, pp. 41–45.
	
	\bibitem{dsent}
	C. Sun et al., "DSENT - a tool connecting emerging photonics with electronics for opto-electronic networks-on-chip modeling," in Proc. 6th Int'l Symp. Networks-on-Chip (NoCS), 2012, pp. 201-210.
	
\end{thebibliography}
	%
	
\end{document}